\documentclass[runningheads]{llncs}

\usepackage{eccv}

\usepackage{eccvabbrv}

\usepackage{graphicx}
\usepackage{booktabs}

\usepackage[accsupp]{axessibility}  % Improves PDF readability for those with disabilities.

\usepackage{hyperref}
\begin{document}

\title{ProtoFair: Fair Self-Supervised Contrastive Learning via Pseudo-Counterfactual Pairs} 

\titlerunning{ProtoFair}

% Second Author\inst{2,3}\orcidlink{1111-2222-3333-4444} \and
% Third Author\inst{3}\orcidlink{2222--3333-4444-5555}}

\author{Marah Halawa \and Olaf Hellwich}

\authorrunning{M.~Halawa et al.}

% Springer Heidelberg, Tiergartenstr.~17, 69121 Heidelberg, Germany

\institute{Technische Universität Berlin \\
\email{halawa@campus.tu-berlin.de},
\email{olaf.hellwich@tu-berlin.de}}

\maketitle

\begin{abstract}

Self-supervised learning methods learn high-quality visual representations, yet recent studies show that these representations often capture demographic biases present in the training data. Existing fairness-aware methods address this by redesigning the self-supervised objective itself, limiting portability across the rapidly evolving landscape of self-supervised learning (SSL) frameworks. We propose ProtoFair, a fairness-aware contrastive loss designed to work alongside existing SSL objectives without modifying them. ProtoFair leverages unsupervised prototype clustering to identify pseudo-counterfactual pairs: samples sharing the same cluster assignment but belonging to different sensitive groups. By pulling these content-matched, cross-group samples together in the embedding space, ProtoFair encourages the encoder to learn representations that are invariant to the sensitive attribute. The method requires only sensitive attribute annotations, no target labels, and integrates seamlessly with both SimCLR and SupCon. Experiments on CelebA and UTKFace demonstrate consistent fairness improvements while maintaining competitive accuracy.
  % \keywords{First keyword \and Second keyword \and Third keyword}
\end{abstract}

\section{Introduction}
\label{sec:intro}
Self-supervised contrastive learning has emerged as a leading approach for learning visual representations without requiring manual annotation. Methods such as SimCLR~\cite{chen2020simple}, MoCo~\cite{he2020momentum}, and SupCon~\cite{khosla2020supervised} have demonstrated that contrastive objectives can produce representations that compete with their supervised counterparts across a range of downstream tasks. However, recent work has revealed that self-supervised representations are not immune to societal biases. Models trained without explicit labels still encode demographic information present in the training data ~\cite{Sirotkin_2022_CVPR}, ~\cite{steed2021image} leading to unfair outcomes in downstream applications such as facial attribute classification.

Addressing fairness in representation learning has attracted growing attention. 
Two prominent lines of work have emerged. The first focuses on adversarial debiasing methods~\cite{kim2019learning}, which trains an auxiliary discriminator to remove sensitive information from representations. The second develops modified contrastive objectives~\cite{park2022fair, zhang2023fairness}, which redesign the loss function to reduce bias. While effective, both strategies share a common limitation: they require replacing or fundamentally altering the base training objective. Adversarial methods introduce minimax optimization known for its instability, while fairness-aware contrastive losses such as FSCL~\cite{park2022fair} modify the negative sampling strategy and require target task labels. As self-supervised learning continues to evolve rapidly, with new methods regularly achieving state-of-the-art performance, fairness approaches tightly coupled to a specific loss function must be redesigned for each new method.

In this paper, we ask a different question: rather than redesigning SSL objectives for fairness, can we introduce a lightweight regularizer that makes \emph{other} existing SSL methods fairer? We propose \textbf{ProtoFair Loss}, a fairness-aware contrastive loss added to the base SSL loss as an auxiliary term, leaving the original objective entirely unchanged. ProtoFair identifies pseudo-counterfactual pairs, which are samples that share similar semantic content but belong to different demographic groups, and encourages their representations to be similar. The key conceptual motivation comes from counterfactual fairness~\cite{kusner2017counterfactual}: a representation is fair if it would remain unchanged were an individual's sensitive attribute different. By pulling such pairs together, ProtoFair actively encourages the learned representation to reflect content rather than demographic membership.

  \paragraph{The main contributions of this paper are summarized as follows}
  \begin{enumerate}
      \item A \textbf{plug-in fairness-aware contrastive loss} that can
      be added to other SSL objective without
      modifying them, requiring only sensitive attribute annotations,  not target task labels.

      \item A \textbf{pseudo-counterfactual pair construction strategy}, where samples from different sensitive groups assigned to the same cluster are treated as positives in a  contrastive objective, actively encouraging group-invariant representations. To the best of our knowledge, \textit{this is the first work to} (i) employ unsupervised clustering as a mechanism for fairness in contrastive learning and (ii) construct pseudo-counterfactual pairs through cluster assignments.

    % \item A novel connection between \textbf{clustering-based SSL and fairness}: we use unsupervised cluster assignments as a proxy for semantic content, enabling the construction of cross-group counterfactual pairs without target labels. While clustering has been widely used in SSL for improving representation quality, to our knowledge we are the first to repurpose it for fairness.

    %  \item A \textbf{pseudo-counterfactual pair definition}
    %   for contrastive fairness: samples assigned to the same
    %   cluster but belonging to different sensitive groups are
    %   treated as positives, actively encouraging
    %   group-invariant representations for semantically
    %   similar content.

      \item \textbf{Empirical validation} on
      CelebA~\cite{liu2015faceattributes}, 
      UTKFace~\cite{zhifei2017cvpr}, and NIH Chest X-rays ~\cite{8099852} showing improved
      equalized odds~\cite{hardt2016equality} across multiple
      base SSL methods while maintaining competitive accuracy.
  \end{enumerate}

\section{Related Work}
\label{sec:related}
         
  \subsection{Self-Supervised Learning}                                                         
  Self-supervised learning (SSL) has appeared as a powerful paradigm for learning
  visual representations without requiring annotation. 
  Among the notable methods in this domain is contrastive learning, which learns representations by pulling similar (positive) pairs together while pushing dissimilar (negative) pairs apart in the embedding space. 
  SimCLR~\cite{chen2020simple}
  established a framework for self-supervised contrastive
  learning by treating two augmented views of the same image as positives and all
  other samples in the batch as negatives. Leveraging this concept, Khosla et
  al.~\cite{khosla2020supervised} generalized the contrastive loss to the
  supervised setting with SupCon, which leverages label information to treat all
  samples sharing the same class as positives, achieving superior performance over
  the standard cross-entropy loss on several benchmarks.
One significant challenge in contrastive learning is obtaining sufficient negative samples for effective training. He et al.~\cite{he2020momentum}
  addressed this through Momentum Contrast (MoCo), which maintains a dynamic queue of representations encoded by a momentum-updated encoder, decoupling dictionary size from mini-batch size and enabling a large set of negatives, without needing enormous batches. The momentum encoder and queue mechanism introduced by MoCo have
  become influential design patterns in SSL. In our work, we draw inspiration from MoCo's cross-batch queue to construct our feature queue, which
  stores representations, cluster assignments, and sensitive attribute labels from
  past batches, enabling discovery of pseudo-counterfactual pairs beyond the
  current mini-batch.

While contrastive methods focus on distinguishing individual instances, clustering based approaches aim to capture higher-level semantic structure in self-supervised representation learning.
  DeepCluster~\cite{caron2018deep} pioneered this direction by alternating between
  clustering features with K-Means and using the resulting cluster assignments as
  pseudo-labels for discriminative training. However, this offline alternation can
  be computationally expensive. SwAV~\cite{caron2020unsupervised}
  addressed this by using an online clustering approach that enforces consistency between cluster assignments of different augmented views of the same image. This eliminates the need for explicit pairwise comparisons and enabling scalable training. Prototypical Contrastive Learning
  (PCL)~\cite{li2021prototypical} further advanced clustering-based SSL by
  introducing momentum-updated prototypes as latent cluster centroids, using an
  expectation-maximization framework where prototypes concentrate representations
  around semantically meaningful modes. Our ProtoFair method builds directly on
  the momentum prototype mechanism of PCL: we maintain a set of $K$ prototypes
  updated via exponential moving average (EMA) that define cluster
  assignments over feature representations. However, while PCL and previously mentioned methods use clustering exclusively for improving representation quality, we
  repurpose the cluster structure for a fundamentally different goal, namely fairness.
  % To our knowledge, clustering-based pseudo-labels have not previously been exploited for constructing fairness-aware contrastive objectives in SSL.
  While clustering-based pseudo-labels have been widely adopted in SSL~\cite{caron2020unsupervised,
  li2021prototypical}, their use for constructing fairness-aware contrastive objectives remains largely unexplored.

  Despite the success of self-supervised methods in representation learning for multiple downstream tasks, recent studies have shown that these methods can encode societal biases present in the training data. Steed and
  Caliskan~\cite{steed2021image} demonstrated that the examined self-supervised models
  encode demographic biases. Although the representations are learned without explicit labels, they continue to capture misleading correlations between visual features and sensitive attributes.
Additionally, Sirotkin et al.~\cite{Sirotkin_2022_CVPR} examined three types of self-supervised learning models, including contrastive, geometric, and clustering-based, and found that contrastive models tend to inherit more biases than other SSL approaches.
  This motivates the need for fairness-aware self-supervised methods.

  \subsection{Fairness in Representation Learning}

  Fairness in machine learning has been extensively investigated, particularly within supervised learning settings. Hardt et al.~\cite{hardt2016equality} formalized the notion of equalized odds, requiring that a classifier’s true positive and false positive rates be equal across demographic groups. This approach has emerged as a widely accepted fairness standard. Accordingly, we use equalized odds as our main evaluation metric to assess fairness in downstream classification tasks.

  Early approaches to fair representation learning relied on adversarial debiasing. Kim et al.~\cite{kim2019learning} proposed ``Learning Not to Learn'' which employs an adversarial training objective that penalizes the
  encoder for producing representations from which sensitive attributes can be predicted. Despite their effectiveness, training adversarial methods often faces instability, and require careful hyperparameter tuning to balance the
  minimax objectives~\cite{kim2019learning}. Moreover, adversarial approaches
  typically operate as modifications to the training procedure itself, replacing
  or fundamentally altering the base loss rather than complementing it.

  In the context of face recognition and facial attribute classification, several works have addressed fairness from different angles. Karkkainen and
  Joo~\cite{karkkainen2021fairface} introduced the FairFace dataset and highlighted the racial biases present in existing face recognition systems.
  Park et al.~\cite{park2022fair} proposed FSCL (Fair Supervised Contrastive Learning), which achieves fairness 
  by modifying the selection of negative samples in the supervised contrastive loss so that sensitive attribute information is not exploited.
  % by restricting the denominator of the
  % contrastive loss to same-group samples only, implementing what can be characterized as \emph{passive} fairness.
  % Preventing the loss from explicitly contrasting samples across demographic groups. 
  While FSCL represents an
  important step toward fairness-aware contrastive learning, it operates by
  constraining what the model learns \emph{not} to distinguish, rather than
  actively encouraging group-invariant representations for similar content.
  Furthermore, FSCL requires target task labels to define positive pairs
  within the supervised contrastive framework. Our ProtoFair loss is
  complementary to FSCL and can be combined with it. FSCL provides fairness through restriction to the denominator of supervised contrastive, while our loss provides fairness by explicitly pulling cross-group pairs together.

  Several other works have tackled fairness in facial attribute classification.
  Chiu et al.~\cite{chiu2023fair} proposed a fair multi-exit framework which uses intermediate classifiers to produce accurate yet less biased predictions. Park et
  al.~\cite{park2021learning} introduced a disentangled representation approach
  for fair facial attribute classification via fairness-aware information
  alignment, which separates sensitive attribute information from task-relevant
  features.
  %   The idea of disentangling attribute-specific factors in facial
  % representations has been explored in related contexts; for instance,
  % Halawa et al.~\cite{halawa2020disentangled} proposed learning
  % disentangled expression representations from facial images, separating
  % identity from expression factors. While our work operates in a
  % different setting---encouraging demographic-group invariance for
  % content-matched samples rather than explicit factor
  % disentanglement---both approaches share the broader motivation that
  % facial representations should not conflate semantically distinct
  % sources of variation.
  In other work ~\cite{dinca2024improving} they explored vision-language driven
  image augmentation to improve fairness by generating augmented training samples
  that balance demographic representation.
In the broader fairness-aware contrastive learning literature, Chai and
  Wang~\cite{chai2022self} investigated self-supervised fair representation
  learning and demonstrated that standard self-supervised objectives do not inherently provide fairness guarantees, when trained alone without access to demographic labels. Zhang et al.~\cite{zhang2023fairness} proposed
  fairness-aware contrastive learning methods that modify the contrastive
  objective to reduce bias, especially when sensitive attribute labels are only partially available. Primarily through re-weighting and  augmentation strategies tied to sensitive attributes.
%%%%%%%%% this can be moved to introduction %%%%%%%%%%%%%%%%%%%
A critical distinction of our approach is its role as a \emph{plug-in regularizer}. Existing fairness methods for contrastive learning typically replace the base loss (e.g., adversarial debiasing ~\cite{kim2019learning}), require target labels (e.g., FSCL ~\cite{park2022fair}). In contrast, \texttt{ProtoFair} is an additive regularization term that can be appended to any self-supervised objective without modifying it. This plug-and-play design preserves representation quality while improving fairness, requiring only sensitive attribute annotations rather than target task labels, and is thus suitable for fully self-supervised settings.
%%%%%%%%%%%%%%%%%%%%%%%%%%%%%%%%%%%%%%%%%%%%%%%%%%%%%%%%%%%%%%%%%%%

Our method draws conceptual motivation from the counterfactual fairness framework of Kusner et al.~\cite{kusner2017counterfactual}, which defines a decision as fair if it would remain unchanged were an individual's sensitive attribute different, relying on structural causal models to reason about interventions while holding causally upstream variables fixed. We implement this intuition by constructing \emph{pseudo-counterfactual pairs}: samples from different sensitive groups assigned to the same unsupervised cluster, which serves as a proxy for shared semantic content. Unlike true counterfactuals, our approach does not require knowledge of the causal graph; instead, it approximates the counterfactual condition under the assumption that cluster assignments capture task-relevant, non-sensitive factors. While this assumption is weaker than full causal identification, our empirical results demonstrate it is sufficient to yield substantial fairness improvements in practice.
  
  % The key
  % advantage of our approach is that it operationalizes the intuition of
  % counterfactual fairness within a contrastive learning framework without
  % requiring a causal model specification, making it practically applicable to
  % high-dimensional visual data where causal graphs are typically unknown.

\section{Methodology}
\label{sec:method}
We propose \textbf{ProtoFair Loss}, a fairness-aware contrastive regularizer that can be incorporated into existing self-supervised learning frameworks to promote representations invariant to sensitive attributes. This ensures samples with similar semantic content are embedded close to one another, regardless of their demographic group membership. The core idea is to leverage unsupervised cluster assignments as pseudo-labels for semantic content, enabling the construction of \emph{pseudo-counterfactual pairs}, i.e., samples that share similar content but differ in sensitive group membership, without requiring task-specific labels. Only sensitive attribute annotations are required. Concretely, ProtoFair Loss pulls together representations of samples that are semantically similar (assigned to the same cluster) but belong to different demographic groups, encouraging the model to learn features invariant to the sensitive attribute within each content cluster. 

  \begin{figure}[tb]                                                                                                                      
    \centering                                                                                                                            
    \includegraphics[width=\textwidth]{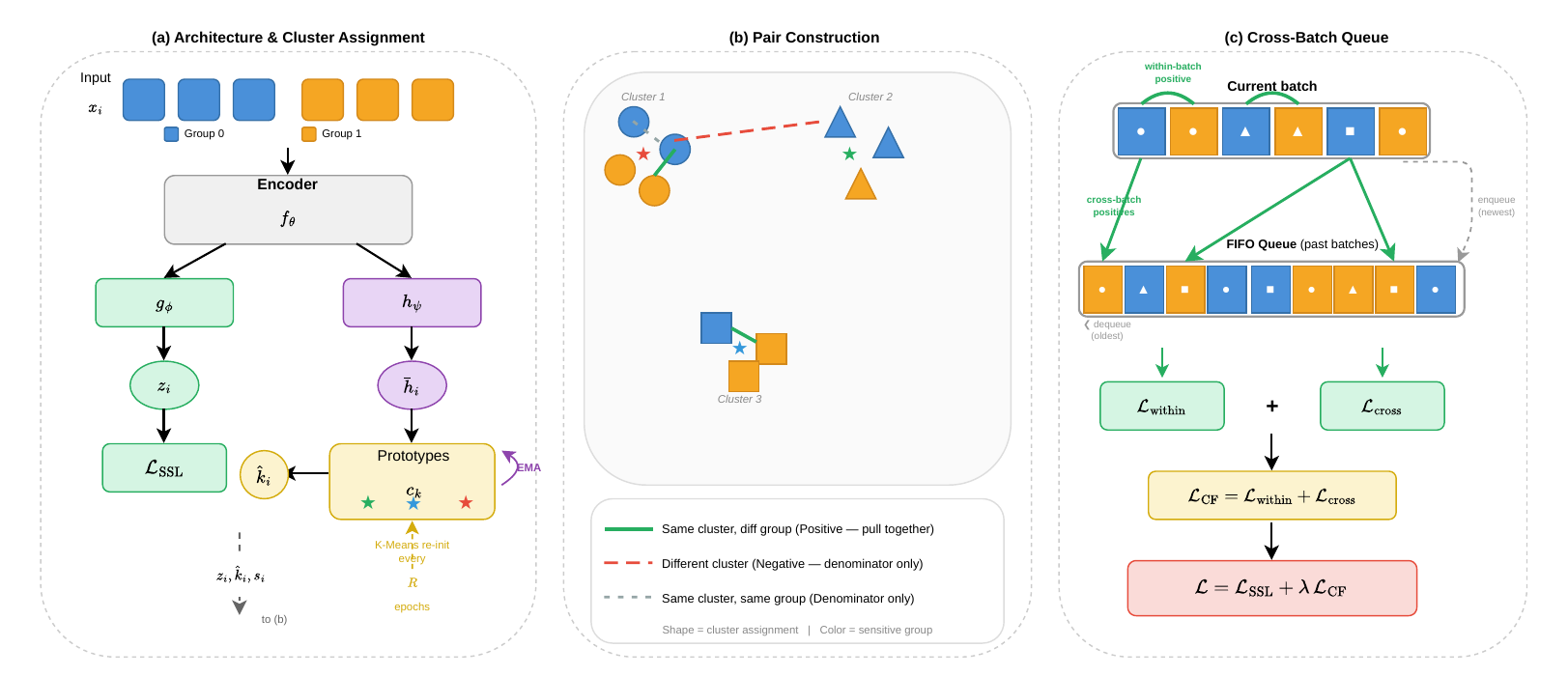}                                                                                  
    \caption{Illustration of the key steps in the ProtoFair loss.                                                                                    
      \textbf{(a)}~A shared encoder $f_\theta$ produces representations projected by two heads: a contrastive head
      $g_\phi$ (\emph{green}) for the base SSL loss and a cluster head
      $h_\psi$ (\emph{purple}) for computing cluster assignments via
      momentum-updated prototypes (\emph{stars}). Prototypes are
      initialized with K-Means and tracked between re-initializations
      using exponential moving average updates.
      \textbf{(b)}~Pseudo-counterfactual pair construction in the
      embedding space. Samples are colored by sensitive group and shaped
      by cluster assignment. Pairs sharing the same cluster but from different groups (green solid lines) are positives, whereas different-cluster pairs (red dashed lines) and same-cluster same-group pairs (gray dotted lines) are non-positives, appear
      only in the denominator.
      \textbf{(c)}~Within-batch positives are identified in the current
      mini-batch, while cross-batch positives are discovered by matching
      against a FIFO queue of past batch representations. The two
      components form the ProtoFair regularizer
      $\mathcal{L}_{\text{CF}}$, which is added to the base SSL loss
      $\mathcal{L}_{\text{SSL}}$.}
    \label{fig:protofair_overview}
  \end{figure}

% The loss consists of two complementary components:
%   %
%   \begin{enumerate}
%       \item \textbf{Within-batch loss}---a fairness contrastive loss
%       computed over pseudo-counterfactual pairs found in the current
%       mini-batch.
%       \item \textbf{Cross-batch loss}---the same principle applied by
%       matching current-batch samples against a queue of past batches,
%       expanding the opportunity to discover cross-group pairs,
%       particularly when batches are small or sensitive groups are
%       imbalanced.
%   \end{enumerate}
  
\subsection{Problem Formulation}

  Let $f_\theta$ denote an encoder network parameterized by $\theta$,
  and let $g_\phi$ denote a projection head that maps encoder outputs
  to a normalized embedding space. For an input sample $x_i$, we obtain
  the \mbox{L2-normalized} representation
  $z_i = g_\phi(f_\theta(x_i)) \in \mathbb{R}^d$. Each sample is
  associated with a sensitive attribute $s_i \in \{0, 1\}$ (e.g.,
  gender), which is assumed to be known during training.

Existing self-supervised methods such as
  SimCLR~\cite{chen2020simple}, SupCon~\cite{khosla2020supervised}
  % ,
  % and their fairness-aware extensions like FSCL~\cite{park2022fair}
  have demonstrated strong representation learning capabilities.
  However, incorporating fairness typically requires redesigning the
  contrastive objective itself, limiting portability across SSL
  frameworks and often necessitating costly retraining from scratch.
We take a different approach: rather than replacing the SSL
  objective, we introduce ProtoFair Loss as an auxiliary regularizer
  $\mathcal{L}_{\text{CF}}$ that is added to an existing SSL loss
  $\mathcal{L}_{\text{SSL}}$:
  \begin{equation}
      \mathcal{L} = \mathcal{L}_{\text{SSL}}
                  + \lambda \, \mathcal{L}_{\text{CF}}
      \label{eq:total_loss}
  \end{equation}
  %
  % where $\lambda > 0$ controls the fairness-utility trade-off.
where $\lambda > 0$ controls the strength of the fairness regularization. The base loss $\mathcal{L}_{\text{SSL}}$ can be any self-supervised objective (e.g., SimCLR, SupCon) and remains
  \emph{unchanged}. This plug-in design offers two practical
  advantages: (i)~it avoids modifying existing SSL methods that
  already produce high-quality representations, and (ii)~it enables
  fairness regularization to benefit from ongoing advances in SSL
  without requiring method-specific adaptations. Beyond the projection head $g_\phi$ used by the base SSL loss, our method introduces a separate \emph{cluster projection
  head} $h_\psi: \mathbb{R}^m \rightarrow \mathbb{R}^d$ that maps encoder representations to a clustering embedding space. This head operates on the same encoder output as $g_\phi$ but is updated only through the ProtoFair loss, decoupling the base SSL and the fairness objectives. The cluster head output is used exclusively by the ProtoFair regularizer for computing cluster assignments, as described in the following subsection.

 \subsection{ProtoFair: Cluster-Fair Contrastive Loss}
% The core principle of our loss is to pull together samples that share the same cluster assignment but differ in their sensitive attribute, thereby encouraging the encoder to produce similar representations for semantically similar content regardless of the sensitive group.

% \textbf{Positive pair definition} A pair $(i, j)$ is considered positive if and only if:
% \begin{itemize}
%    \item They belong to the same cluster: $k_i = k_j$, and
%    \item They belong to different sensitive groups: $s_i \neq s_j$.
% \end{itemize}

% These pairs serve as pseudo-counterfactuals: samples with similar semantic content but different sensitive attributes. By pulling them together, we encourage the representation space to be invariant to the sensitive attribute for samples with shared content.

\subsubsection{Momentum-Updated Cluster Prototypes} 
  To construct pseudo counterfactual pairs without target labels,
  we require a notion of semantic content similarity. We obtain
  this through a set of $K$ prototype vectors
  $\{c_k\}_{k=1}^{K} \subset \mathbb{R}^d$, maintained as
  non-learnable running estimates in the cluster embedding space
  produced by $h_\psi$. The prototypes are not updated via
  backpropagation; instead, they are maintained through K-Means
  initialization and momentum-based tracking, as described below.

  \paragraph{Initialization.}
 After a warmup period of several epochs, during which the encoder is trained using only the base SSL loss, the prototypes are initialized by performing K-means clustering over the full set of training representations in cosine space:
  \begin{equation}
      \{c_k\}_{k=1}^{K} \leftarrow
      \texttt{K\text{-}Means}\!\Big(\big\{
          \bar{h}_i = \tfrac{h_\psi(f_\theta(x_i))}
          {\|h_\psi(f_\theta(x_i))\|}
      \big\}_{i=1}^{N}\Big).
      \label{eq:kmeans_init}
  \end{equation}
  To prevent prototype drift as the representation space evolves
  during training, K-Means re-initialization is performed
  periodically every $R$ epochs over the full training set.

  \paragraph{Momentum update.}
  Between re-initializations, prototypes are updated every
  training iteration via exponential moving average (EMA) to
  smoothly track the evolving feature space. Given a mini-batch
  of cluster-head features $\{\bar{h}_i\}_{i=1}^{B}$ with hard
  assignments
  $\hat{k}_i = \arg\max_k \, \bar{h}_i^\top c_k$, the update
  for prototype $c_k$ is:
  \begin{equation}
      c_k \leftarrow \text{normalize}\!\Big(
          m \cdot c_k + (1 - m) \cdot
          \frac{\sum_{i:\hat{k}_i=k} \bar{h}_i}
               {|\{i : \hat{k}_i = k\}|}
      \Big),
      \label{eq:proto_ema}
  \end{equation}
  where $m \in [0,1)$ is the momentum coefficient. This
  continuous tracking ensures that cluster assignments remain
  meaningful as the encoder improves, without the cost of
  running full-dataset K-Means at every iteration.

  \paragraph{Cluster assignments.}                                                                                             
  Each sample is assigned to its nearest prototype in the                                                                      
  cluster embedding space based on cosine similarity:                                                                          
  \begin{equation}
      \hat{k}_i = \arg\max_k \; \bar{h}_i^\top c_k
      \label{eq:cluster_assignment}
  \end{equation}
  These hard assignments serve as pseudo-content labels for
  constructing counterfactual pairs, as described next.

  \paragraph{Gradient flow.}
  The cluster assignments are detached from the computational graph when passed to the ProtoFair loss.
   This design follows an EM-style alternating optimisation: the cluster assignments act as a fixed E-step, providing
   pseudo-labels that define which pairs serve as positives, while the fairness contrastive update acts as the 
  M-step, optimising the encoder and projection head, mirroring PCL~\cite{li2021prototypical}.
  Detachment is essential because, without it, gradients from $\mathcal{L}{\text{CF}}$ could manipulate the cluster
  assignments themselves rather than improving the representations. In the degenerate case, the model could collapse
  all samples into a single cluster, trivially satisfying the same-cluster criterion without producing fairer
  representations. By treating cluster assignments as fixed pseudo-labels, the clustering module reflects the underlying content similarity and is neither rewarded nor penalised by the fairness objective. The fairness loss
  $\mathcal{L}{\text{CF}}$ therefore updates only the encoder $f_\theta$ and the contrastive projection head $g_\phi$
   through the feature similarity computation in its contrastive objective. The cluster head $h_\psi$ is not directly
   updated by $\mathcal{L}{\text{CF}}$; it evolves indirectly as the shared encoder $f\theta$ is updated by both
  $\mathcal{L}{\text{SSL}}$ and $\mathcal{L}{\text{CF}}$, as defined in Eq.~\eqref{eq:total_loss}.
  % The cluster assignments are \emph{detached} from the
  % computational graph when passed to the ProtoFair loss.
  % The fairness loss $\mathcal{L}_{\text{CF}}$ updates the
  % encoder $f_\theta$ and the contrastive projection head
  % $g_\phi$ through the feature similarity computation in
  % its contrastive objective. The cluster head $h_\psi$ is
  % not directly updated by $\mathcal{L}_{\text{CF}}$; it
  % evolves indirectly as the shared encoder $f_\theta$ is
  % updated by both $\mathcal{L}_{\text{SSL}}$ and
  % $\mathcal{L}_{\text{CF}}$.

    \subsubsection{Pseudo-Counterfactual Pair Construction}

  The central idea of ProtoFair is to identify pairs of samples
  that share semantic content but differ in their sensitive
  attribute, approximating counterfactual pairs without requiring target labels. Given a mini-batch of $B$ samples with features
  $\{z_i\}_{i=1}^{B}$, hard cluster assignments
  $\{\hat{k}_i\}_{i=1}^{B}$, and sensitive attributes
  $\{s_i\}_{i=1}^{B}$, we define the \emph{pseudo-counterfactual
  positive set} for sample $i$ as:
  \begin{equation}
      \mathcal{P}_i = \big\{\, j \neq i \;\big|\;
          \hat{k}_j = \hat{k}_i \;\;\text{and}\;\;
          s_j \neq s_i \,\big\}
      \label{eq:positive_set}
  \end{equation}
  That is, sample $j$ is a positive for sample $i$ if and only
  if they belong to the same cluster (proxy for shared content)
  and to different sensitive groups. This yields three categories
  of pairs:
  \begin{itemize}
      \item \textbf{Same cluster, different sensitive group}
      $(\hat{k}_j = \hat{k}_i,\; s_j \neq s_i)$:
      pseudo-counterfactual pairs (\emph{pulled together}).
      % (positive, in numerator).
      \item \textbf{Different cluster}
      $(\hat{k}_j \neq \hat{k}_i)$:
      semantically dissimilar pairs (\emph{pushed apart}).
      % (denominator only).
      \item \textbf{Same cluster, same sensitive  group}
      $(\hat{k}_j = \hat{k}_i,\; s_j = s_i)$:
      same-content, same-group pairs (\emph{pushed apart}).
      % (denominator only, excluded from numerator).
  \end{itemize}
  The third category is key: by excluding same-group pairs from the numerator even when they share the same cluster, the loss specifically targets cross-group alignment rather than general within-cluster similarity.

  \subsubsection{Within-Batch Contrastive Fairness Loss}

  Given the positive set $\mathcal{P}_i$ defined above, we
  formulate the within-batch fairness loss following the
  contrastive learning framework. For each sample $i$ with
  $|\mathcal{P}_i| > 0$, the loss is:
  \begin{equation}
      \mathcal{L}_{\text{within}} =
      -\frac{1}{|\mathcal{V}|} \sum_{i \in \mathcal{V}}
      \frac{1}{|\mathcal{P}_i|} \sum_{j \in \mathcal{P}_i}
      \log \frac{
          \exp\!\big(\text{sim}(z_i, z_j) / \tau\big)
      }{
          \displaystyle\sum_{\substack{k=1 \\ k \neq i}}^{B}
          \exp\!\big(\text{sim}(z_i, z_k) / \tau\big)
      }
      \label{eq:within_loss}
  \end{equation}
  where $\text{sim}(z_i, z_j) = z_i^\top z_j$ is the cosine
  similarity between L2-normalized features, $\tau$ is the
  temperature, and $\mathcal{V} = \{i : |\mathcal{P}_i| > 0\}$
  is the set of samples that have at least one
  pseudo-counterfactual partner in the batch. The denominator sums over \emph{all} non-self pairs in the
  batch, following the standard contrastive denominator. 
  % This
  % ensures that pseudo-counterfactual pairs are pulled together
  % relative to the full set of negatives, maintaining the
  % discriminative power of the representation.

  % Same-cluster same-group pairs are excluded from the numerator
  % (and thus appear only in the denominator) because they carry
  % no cross-group information. Including them would encourage
  % general cluster cohesion---a task already handled by the base
  % SSL loss---rather than the cross-group alignment that is the
  % specific goal of the fairness regularizer.

\subsubsection{Cross-Batch Queue for Expanded Pair Discovery}

  A practical limitation of the within-batch loss is that
  pseudo-counterfactual pairs require samples from the same
  cluster but different sensitive groups to co-occur in a
  mini-batch. When batches are small or sensitive groups are
  imbalanced, such pairs may be scarce, weakening the
  fairness signal. To address this, we maintain a first-in-first-out (FIFO)
  queue $\mathcal{Q}$ that stores representations from the
  $M$ most recent batches. For each past sample, the queue retains a tuple of its feature vector, cluster assignment, and sensitive attribute: $\mathcal{Q} = \{(z_j^q, \, \hat{k}_j^q, \,
  s_j^q)\}_{j=1}^{Q}$, where $Q = M \times B$ is the total
  queue capacity. Queue entries are stored as detached
  tensors and are not involved in gradient computation.
  % the queue retains a tuple of its feature vector, soft cluster
  % probabilities, and sensitive attribute:
  % $\mathcal{Q} = \{(z_j^q, \, q_j^q, \, s_j^q)\}_{j=1}^{Q}$,
  % where $Q = M \times B$ is the total queue capacity. Queue
  % entries are stored as detached tensors and are not involved
  % in gradient computation.

  At each training step, current-batch samples are matched
  against the queue to find additional cross-group pairs. The
  positive set for sample $i$ with respect to the queue is:
  \begin{equation}
      \mathcal{P}_i^q = \big\{\, j \in \mathcal{Q} \;\big|\;
          \hat{k}_j^q = \hat{k}_i \;\;\text{and}\;\;
          s_j^q \neq s_i \,\big\}
      \label{eq:queue_positive_set}
  \end{equation}
  and the cross-batch loss follows the same contrastive
  formulation:
  \begin{equation}
      \mathcal{L}_{\text{cross}} =
      -\frac{1}{|\mathcal{V}^q|} \sum_{i \in \mathcal{V}^q}
      \frac{1}{|\mathcal{P}_i^q|}
      \sum_{j \in \mathcal{P}_i^q}
      \log \frac{
          \exp\!\big(\text{sim}(z_i, z_j^q) / \tau\big)
      }{
          \displaystyle\sum_{k \in \mathcal{Q}}
          \exp\!\big(\text{sim}(z_i, z_k^q) / \tau\big)
      }
      \label{eq:cross_loss}
  \end{equation}
  where $\mathcal{V}^q = \{i : |\mathcal{P}_i^q| > 0\}$.
  Note that the denominator sums over all queue entries,
  providing a large and diverse set of negatives. After each
  forward pass, the queue is updated by enqueuing the current
  batch and dequeuing the oldest entries.

  \subsubsection{Full Loss Formulation}

  The complete ProtoFair regularizer combines the within-batch
  and cross-batch components:
  \begin{equation}
      \mathcal{L}_{\text{CF}} =
      \mathcal{L}_{\text{within}} +
      \mathcal{L}_{\text{cross}}
      \label{eq:protofair_loss}
  \end{equation}
  This is combined with the base SSL objective as defined in
  Eq.~\eqref{eq:total_loss}:
  \begin{equation*}
      \mathcal{L} = \mathcal{L}_{\text{SSL}}
                  + \lambda \, \mathcal{L}_{\text{CF}}
  \end{equation*}
where $\lambda > 0$ determines the magnitude of the fairness regularization effect. The ProtoFair regularizer is not applied from the beginning
  of training. The model first trains for several warmup epochs using only the base SSL loss
  $\mathcal{L}_{\text{SSL}}$, allowing the encoder to learn
  meaningful representations before prototypes are initialized
  and the fairness loss is activated. The
  queue is initially empty and populates naturally over time. The cross-batch loss contributes zero
  until the queue contains entries.

\section{Experiments}
\label{sec:experiments}
\subsection{Datasets}

We evaluate ProtoFair on two widely-used facial-attribute benchmarks (CelebA, UTKFace) and one medical-imaging benchmark (NIH Chest X-rays).
  
  \textbf{CelebA~\cite{liu2015faceattributes}} contains over 200K face images annotated with 40 binary attributes. We use the standard train/val/test partition and follow~\cite{park2022fair} for target and sensitive attribute notation in all table results.
  % we denote the attributes as:
  % \texttt{a}~(attractiveness), \texttt{b}~(big nose),
  % \texttt{e}~(bags under eyes), \texttt{m}~(male), and \texttt{y}~(young), where
  % $T$ and $S$ refer to target and sensitive attributes,
  % respectively.
  
\textbf{UTKFace~\cite{zhifei2017cvpr}} consists of over 20K face images labeled with age, gender, and ethnicity. Following~\cite{park2022fair}, we binarize 
% age ($<35$ vs.\ $\geq 35$) and
ethnicity (Caucasian vs.\ non-Caucasian). We construct training sets with varying imbalance ratios $\alpha \in \{2, 3, 4\}$, where $\alpha$ controls the skew between demographic groups. Val/test sets are kept balanced.

\textbf{NIH Chest X-rays ~\cite{8099852}} is a large-scale medical imaging benchmark of 112{,}120  chest radiographs from 30{,}805 patients, each annotated with up to 14 thoracic pathologies.

  \subsection{Implementation Details}

  \paragraph{Encoder training, and downstream evaluation.}
  We use ResNet-18~\cite{He2015DeepRL} as the backbone encoder
  with two separate MLP projection heads: a contrastive head
  % $g_\phi$ (output dimension 128) 
  for the base SSL loss, and
  a cluster head 
  % $h_\psi$ (output dimension 128) 
  for prototype
  computation. Training uses SGD with momentum 0.9, weight
  decay $10^{-4}$, and an initial learning rate of 0.1 with
  cosine annealing. Following the linear evaluation protocol
  of~\cite{park2022fair}, we freeze the trained encoder and
  train a linear classifier on top of the encoder features
  using cross-entropy loss. 
  % Input images are resized to $128 \times 128$.
  % with standard augmentations: random resized crop
  % (scale 0.2--1.0), horizontal flip, color jitter, and random
  % grayscale. The batch size is 128. 
  % and training runs for 100 epochs.

  % \paragraph{ProtoFair hyperparameters.}
  % We use $K=10$ cluster prototypes with momentum $m=0.9$ and
  % cluster temperature $\tau_c=0.1$. Prototypes are initialized
  % via K-Means after a warmup of \texttt{warmup\_epochs} epochs
  % and re-initialized every $R=5$ epochs. 
  % The fairness loss is activated at epoch \texttt{start\_epoch}=100 with weight
  % $\lambda=0.3$.
  % and temperature $\tau=0.07$. 
  % The cross-batch queue stores 4 past batches.
  % \paragraph{Downstream evaluation.}
  % The classifier is trained for 10 epochs with SGD.
  % (learning rate 0.1, momentum 0.9, 
  % no weight decay) and batch size 128.
  \paragraph{Evaluation metrics.}
  Following~\cite{park2022fair}, we report classification accuracy (ACC) on the target attribute and measure the degree of equalized odds (EO)~\cite{hardt2016equality}
  across sensitive groups, where lower EO corresponds to fairer predictive outcomes.

  % \begin{itemize}
  %     \item \textbf{Accuracy (ACC)}: Overall top-1
  %     classification accuracy on the target attribute.
  %     \item \textbf{Equalized Odds (EO)}: The average absolute
  %     difference in accuracy across sensitive groups, computed
  %     per target class and averaged:
  %     %
  %     \begin{equation}
  %         \text{EO} = \frac{1}{|\mathcal{C}|}
  %         \sum_{y \in \mathcal{Y}} \sum_{j < k}
  %         \big| \text{Acc}(y, s_j) - \text{Acc}(y, s_k) \big|,
  %         \label{eq:eo}
  %     \end{equation}
  %     %
  %     where $\text{Acc}(y, s_j)$ is the accuracy for samples
  %     with target class $y$ and sensitive group $s_j$, and
  %     $|\mathcal{C}|$ is the total number of group comparisons.
  %     Lower EO indicates better fairness.
  % \end{itemize}

\subsection{Evaluation Results}

Table~\ref{tab:celeba} presents classification results on CelebA across multiple target and sensitive attribute scenarios. To enable direct comparison with existing approaches, we adopt the evaluation protocol and baseline results in ~\cite{park2022fair} and apply ProtoFair on top of SupCon under the same setup. Among the baselines, CE ~\cite{HeZRS16} shows the highest EO violations, confirming that unconstrained training encodes sensitive attribute biases. Prior fairness methods such as FSCL ~\cite{park2022fair} and MFD ~\cite{9578197} improve EO, with varying fairness-accuracy trade-offs. 

  For all SupCon + ProtoFair experiments in Table~\ref{tab:celeba}, we use 10 clusters whose prototypes are           
  initialized via K-Means, with $\lambda{=}0.3$. For the \textit{Bags Under Eyes} and \textit{Big Nose} targets, we apply a    
  100-epoch warmup followed by 5 additional epochs of ProtoFair training. For the \textit{Attractiveness} target with \textit{Male}
   as the sensitive attribute, we keep the 100 epoch warmup but train with ProtoFair for 10 epochs instead of 5. For the
  \textit{Attractiveness} target with \textit{Young} as the sensitive attribute, we start SupCon + ProtoFair  optimization at epoch 10 and
  continue for 90 epochs.
We evaluate the model by freezing the encoder and training a linear classifier on top. 
% for a single epoch. 
% Despite this minimal additional training cost, which amounts to only $5\%$ more epochs than the shared baseline, 
ProtoFair achieves substantial fairness improvements. On (T:b, S:y), EO drops from 16.9 to  6.9 with under one point of accuracy loss. Similarly, On (T:a, S:m), EO drops from 30.5 to  14.2 with accuracy nearly unchanged. On (T:e, S:y), ProtoFair achieves an EO of 1.9, matching the best result of the dedicated fairness method FSCL (1.8) while retaining 1.3 points higher accuracy. 
% ProtoFair lowers EO from 20.7 to 15.8 with a negligible accuracy decrease of 0.4 points.
% Similarly, on (T:e, S:m), EO drops from 20.8 to 10.1 while accuracy remains nearly unchanged at 84.0.
These gains require no adversarial training or supervision beyond the sensitive attribute labels already required by all compared
fairness methods.
% architectural modifications

Notably, SupCon + ProtoFair achieves competitive or superior fairness compared to dedicated fairness methods such as GRL ~\cite{raff2018gradient}, LNL ~\cite{kim2019learning}, and FD-VAE ~\cite{park2021learning}, while consistently maintaining higher classification accuracy.
% It is worth noting that we use a single fixed configuration ($K{=}10$, $\lambda{=}0.3$) across all attribute pairs without per-task tuning, demonstrating ProtoFair's practicality.

  % === TABLE 1: Supervised setting ===
  \begin{table}[tb]
    \caption{Classification results on CelebA. We measure
      classification accuracy (ACC) and equalized odds (EO)                                                                               
      in various target ($T$)--sensitive ($S$) attribute
      scenarios. Lower EO is better. Attributes:                                                                                          
      a=attractiveness, b=big nose, e=bags under eyes, m=male, y=young.}
    \label{tab:celeba}
    \centering
    \scriptsize
    \begin{tabular}{@{}lcccccccccccc@{}}
      \toprule
      & \multicolumn{2}{c}{T:a, S:m}
      & \multicolumn{2}{c}{T:a, S:y}
      & \multicolumn{2}{c}{T:b, S:m}
      & \multicolumn{2}{c}{T:b, S:y}
      & \multicolumn{2}{c}{T:e, S:m}
      & \multicolumn{2}{c}{T:e, S:y} \\
      \cmidrule(lr){2-3} \cmidrule(lr){4-5} \cmidrule(lr){6-7}
      \cmidrule(lr){8-9} \cmidrule(lr){10-11} \cmidrule(lr){12-13}
      Method   & {\scriptsize ACC}  & {\scriptsize EO} 
               & {\scriptsize ACC}  & {\scriptsize EO} 
               & {\scriptsize ACC} & {\scriptsize EO} 
               & {\scriptsize ACC}  & {\scriptsize EO} 
               & {\scriptsize ACC}  & {\scriptsize EO} 
               & {\scriptsize ACC} & {\scriptsize EO}  \\
      \midrule
      % --- Baselines from FSCL paper ---
      % Add baseline rows here
      CE ~\cite{HeZRS16}
      & 79.6 & 27.8 & 79.8 & 16.8 & 84.0 & 17.6
      & 84.5 & 14.7 & 83.9 & 15.0 & 83.8 & 12.7 \\
      GRL ~\cite{raff2018gradient}
      & 77.2 & 24.9 & 74.6 & 14.7 & 82.5 & 14.0
      & 83.3 & 10.0 & 81.9 & 6.7 & 82.3 & 5.9 \\
      LNL ~\cite{kim2019learning}
      & 79.9 & 21.8 & 74.3 & 13.7 & 82.3 & 10.7
      & 82.3 & 6.8 & 81.6 & 5.0 & 80.3 & 3.3 \\
      FD-VAE ~\cite{park2021learning}
      & 76.9 & 15.1 & 77.5 & 14.8 & 81.6 & 11.2
      & 81.7 & 6.7 & 82.6 & 5.7 & 84.0 & 6.2 \\
      MFD ~\cite{9578197}
      & 78.0 & 7.4 & 80.0 & 14.9 & 78.0 & 7.3
      & 78.0 & 5.4& 79.0 & 8.7 & 78.0 & 5.2 \\
      FSCL ~\cite{park2022fair}
      & 79.1 & 11.5 & 79.1 & 13.0 & 82.1 & 7.0
      & 83.8 & 6.4 & 82.7 & 3.8 & 82.0 & 1.8 \\
      \midrule
      SupCon ~\cite{khosla2020supervised}
      & 80.5 & 30.5 & 80.1 & 21.7 & 84.6 & 20.7
      & 84.4 & 16.9 & 84.3 & 20.8 & 84.0 & 10.8 \\
      \midrule
      SupCon + ProtoFair (Ours)
      & 80.3 & 14.2 & 81.5 & 17.7 & 84.2 & 15.8
      & 83.5 & 6.9 & 84.0 & 10.1 & 83.3 & 1.9 \\
      \bottomrule
    \end{tabular}
  \end{table}

%  These results demonstrate that ProtoFair provides meaningful fairness improvements as a lightweight addition to supervised contrastive learning, without
% requiring architectural modifications or adversarial training
  
% While standalone fairness methods such as FSCL and ME-MFD achieve stronger absolute performance on EO, these approaches require architectural modifications (TODO CHECK AGAIN),target labels . In contrast, our proposed regularizer is architecture-agnostic and can be applied to any contrastive self-supervised method with minimal modification—requiring only sensitive attribute annotations and no target task labels. The improvement over the unregularized baseline demonstrates that unsupervised cluster assignments provide a viable proxy for content similarity in constructing cross-group counterfactual pairs

 % This modularity and simplicity represent a different and complementary point in the design space of fair representation learning.
  % === TABLE 2: Incomplete supervised setting ===
                                       
\begin{table}[tb]                                                                                                                       
    \caption{CelebA results without target labels during representation learning, using SimCLR as the base loss. Only sensitive attributes are used. ACC: classification accuracy, EO: equalized odds (lower is fairer). a=attractiveness, m=male.}                                                                                       
    \label{tab:celeba_unsupervised}                                                                                                       
    \centering
    \begin{tabular}{@{}lcc@{}}
      \toprule
      & \multicolumn{2}{c}{T:a, S:m} \\
      \cmidrule(lr){2-3}
      Method & {\scriptsize ACC}  $\uparrow$ & {\scriptsize EO} $\downarrow$ \\
      \midrule
      FSCL$^{*}$ ~\cite{park2022fair}
      & 74.6 & 14.8 \\
      SimCLR ~\cite{chen2020simple} + GRL ~\cite{raff2018gradient}
      & 72.3 & 21.9 \\
      SimCLR ~\cite{chen2020simple}
      & 75.7 & 29.4 \\
      \midrule
      SimCLR + ProtoFair (Ours)
      & 73.3 & 21.9 \\
      \bottomrule
    \end{tabular}
  \end{table}

%  We further evaluate ProtoFair in a self-supervised setting where target labels are unavailable during training. Since ProtoFair does not require target labels by design, it naturally extends to this setting. We evaluate this by combining SimCLR~\cite{chen2020simple} as the base loss with ProtoFair as a fairness regularizer. We follow similar training and evaluation protocol as in Table~\ref{tab:celeba}, replacing SupCon with SimCLR as the base loss. Prototypes are initialized via K-Means using 10 clusters after a 100-epoch warmup, followed by 30 epochs of training with the ProtoFair loss at $\lambda{=}0.3$. The encoder is then frozen and a linear classifier is trained for 2 epochs. We compare against the baseline results reported in~\cite{park2022fair}: SimCLR ~\cite{chen2020simple} , SimCLR with gradient reversal (SimCLR ~\cite{chen2020simple} +GRL ~\cite{raff2018gradient}), and FSCL$^{*}$~\cite{park2022fair}, a variant of FSCL that uses only the augmentation pair as the positive sample, analogous to SimCLR.

% As shown in Table~\ref{tab:celeba_unsupervised}, vanilla SimCLR achieves the highest accuracy (75.7) but also the worst EO (29.4). Adding ProtoFair reduces EO from 29.4 to 21.9, matching SimCLR+GRL while maintaining higher accuracy (73.3 vs.\ 72.3). FSCL$^{*}$ achieves the best fairness (EO of 14.8) owing to its specialized contrastive sampling strategy. Nevertheless, ProtoFair operates as a simple auxiliary loss requiring no architectural modifications, demonstrating its flexibility across both supervised and self-supervised regimes.

We further evaluate ProtoFair in a self-supervised setting where target labels are unavailable during training. Since ProtoFair requires no target labels by design, it naturally extends to this setting. We first combine SimCLR~\cite{chen2020simple} as the base loss with ProtoFair as a fairness regularizer, following the training and evaluation protocol of Table~\ref{tab:celeba} but replacing SupCon with SimCLR. Prototypes are initialized via K-Means with 10 clusters after a 100-epoch warmup, followed by 30 epochs of training with the ProtoFair loss at $\lambda{=}0.3$. The encoder is then frozen and a linear classifier is trained for 2 epochs. We compare against the baselines in~\cite{park2022fair}: SimCLR~\cite{chen2020simple}, SimCLR with gradient reversal (SimCLR+GRL~\cite{raff2018gradient}), and FSCL$^{*}$~\cite{park2022fair}, a variant of FSCL that uses only the augmentation pair as the positive sample, analogous to SimCLR.

As shown in Table~\ref{tab:celeba_unsupervised}, vanilla SimCLR achieves the highest accuracy (75.7) but the worst EO (29.4). Adding ProtoFair reduces EO to 21.9, matching SimCLR+GRL while maintaining higher accuracy (73.3 vs.\ 72.3). FSCL$^{*}$ achieves the best fairness (EO of 14.8) owing to its specialized contrastive sampling strategy. Nevertheless, ProtoFair operates as a simple auxiliary loss requiring no architectural modifications, demonstrating its flexibility across supervised and self-supervised regimes.

To further assess this flexibility, we apply ProtoFair as a regularizer to two additional self-supervised methods, BarlowTwins~\cite{DBLP:conf/icml/ZbontarJMLD21} and BYOL~\cite{BYOL}, using the same setup (10 clusters, 100 epoch warmup). Both converge faster than SimCLR, requiring only 5 to 15 epochs of ProtoFair training. We use $\lambda{=}0.3$ for BarlowTwins ~\cite{DBLP:conf/icml/ZbontarJMLD21} (except $\lambda{=}0.2$ for T:e/S:m) and $\lambda{=}0.2$ for all BYOL~\cite{BYOL} experiments. As shown in Table~\ref{tab:ByBT}, ProtoFair consistently improves EO across both methods and all target/sensitive-attribute pairs, at only a minor accuracy cost. This confirms ProtoFair's compatibility with diverse self-supervised objectives.

\begin{table}[tb]
    \caption{CelebA results without target labels during representation learning, applying ProtoFair to BarlowTwins and BYOL. Only sensitive attributes are used. ACC: classification accuracy, EO: equalized odds (lower is fairer). Attributes: b=big nose, e=bags under eyes, m=male, y=young.}
    \label{tab:ByBT}
    \centering
    % \scriptsize
    % \setlength{\tabcolsep}{3pt}
    \begin{tabular}{@{}lcccccccc@{}}
      \toprule
      & \multicolumn{2}{c}{T:e, S:y}
      & \multicolumn{2}{c}{T:e, S:m}
      & \multicolumn{2}{c}{T:b, S:y}
      & \multicolumn{2}{c}{T:b, S:m} \\
      \cmidrule(lr){2-3} \cmidrule(lr){4-5} \cmidrule(lr){6-7} \cmidrule(lr){8-9}
      Method & ACC $\uparrow$ & EO $\downarrow$& ACC $\uparrow$& EO $\downarrow$& ACC $\uparrow$& EO $\downarrow$& ACC $\uparrow$& EO $\downarrow$\\
      \midrule
      BarlowTwins ~\cite{DBLP:conf/icml/ZbontarJMLD21}
      & 79.87 & 1.23 & 80.03 & 1.64 & 81.71 & 6.72 & 81.72 & 12.07 \\
      BarlowTwins ~\cite{DBLP:conf/icml/ZbontarJMLD21}+ ProtoFair 
      & 80.40 & 1.16 & 79.99 & 1.36 & 80.0 & 1.15 & 80.41 & 5.98 \\
      \midrule
      BYOL ~\cite{BYOL}
      & 82.66 & 11.21 & 82.66 & 17.04 & 81.77 & 8.9 & 81.77 & 13.54 \\
      BYOL ~\cite{BYOL} + ProtoFair 
      & 81.75 & 8.98 & 81.11 & 7.11 & 80.42 & 4.59 & 80.16 & 6.83 \\
      \bottomrule
    \end{tabular}
\end{table}

   \begin{table}[tb]                                                                                                                       
    \caption{Classification results on UTKFace                                                                                            
      (T:~gender, S:~ethnicity) under varying data                                                                                        
      imbalance ratios $\alpha$. ACC: classification                                                                                      
      accuracy, EO: equalized odds.}                                                                                    
    \label{tab:utkface}                                                                                                                   
    \centering    
    \begin{tabular}{@{}lcccccc@{}}
      \toprule
      & \multicolumn{2}{c}{$\alpha=4$}
      & \multicolumn{2}{c}{$\alpha=3$}
      & \multicolumn{2}{c}{$\alpha=2$} \\
      \cmidrule(lr){2-3} \cmidrule(lr){4-5} \cmidrule(lr){6-7}
      Method & {\scriptsize ACC}  $\uparrow$ & {\scriptsize EO} $\downarrow$ & {\scriptsize ACC}  $\uparrow$ & {\scriptsize EO} $\downarrow$ & {\scriptsize ACC} $\uparrow$ & {\scriptsize EO} $\downarrow$\\
      \midrule
      FSCL ~\cite{park2022fair}
      & 90.1 & 2.7 & 92.3 & 1.7 & 91.6 & 1.0 \\
      SupCon ~\cite{khosla2020supervised}
      & 89.8 & 10.6 & 91.6 & 8.4 & 92.0 & 4.5 \\
      % FSCL+
      % & 90.1 & 1.6 & 92.2 & 1.0 & 91.5 & 0.6 \\
      \midrule
      SupCon + ProtoFair (Ours)
      & 89.8 & 6.6 & 90.3 & 4.8 & 90.7 & 2.8 \\
      \bottomrule
    \end{tabular}
  \end{table}

  To assess the robustness of ProtoFair under varying degrees of dataset bias, we evaluate on UTKFace with gender as the target attribute and ethnicity as the sensitive attribute. The imbalance ratio $\alpha$ controls the skew between demographic groups, with higher values   
  indicating stronger bias. We adopt the baseline results reported in~\cite{park2022fair} and apply ProtoFair on top of SupCon. We initialize prototypes via K-Means after a 10-epoch warmup and train with the combined SupCon and ProtoFair loss for the remaining 90 epochs. As described in Section~\ref{sec:method}, prototypes are updated via momentum after each batch and re-initialized with K-Means every 5 epochs to prevent drift from the evolving
  feature space. As shown in Table~\ref{tab:utkface}, ProtoFair consistently reduces EO compared to the SupCon baseline across all imbalance levels. EO decreases from 10.6 to 6.6 at $\alpha=4$, from 8.4 to 4.8 at $\alpha=3$, and from 4.5 to 2.8 at $\alpha=2$, with minimal loss in accuracy. The relative improvement remains stable as bias increases, suggesting that ProtoFair scales gracefully with the degree of imbalance rather than being effective only in low-bias regimes.

  \begin{table}[tb]
  \caption{Results on NIH Chest X-rays ~\cite{8099852} with pneumothorax as the target and patient sex as the sensitive attribute, using SupCon as the base loss.}
  \label{tab:nihcxr}
  \centering
  % \scriptsize
  % \setlength{\tabcolsep}{5pt}
  \begin{tabular}{@{}lcccc@{}}
    \toprule
    Method & ACC $\uparrow$& AUROC $\uparrow$ & AUROC gap $\downarrow$ & EO $\downarrow$ \\
    \midrule
    SupCon ~\cite{khosla2020supervised}           & 89.34 & 0.70 & 0.03 & 1.87 \\
    SupCon + ProtoFair  & 89.28 & \textbf{0.72} & \textbf{0.01} & \textbf{1.79} \\
    \bottomrule
  \end{tabular}
\end{table}

To verify that ProtoFair generalizes beyond facial images, we evaluate it in the medical-imaging domain on NIH Chest X-rays ~\cite{8099852}, using SupCon as the base loss and pneumothorax as the target. As shown in Table~\ref{tab:nihcxr}, ProtoFair reduces the AUROC gap between sex groups (from 0.03 to 0.01) and lowers EO (1.87 to 1.79), while slightly improving overall AUROC (0.70 to 0.72) at negligible accuracy cost. This shows that ProtoFair improves both fairness and AUROC.

   \begin{figure}[tb]
    \centering
    \begin{subfigure}{0.48\linewidth}                                                                                         
      \includegraphics[width=\linewidth]{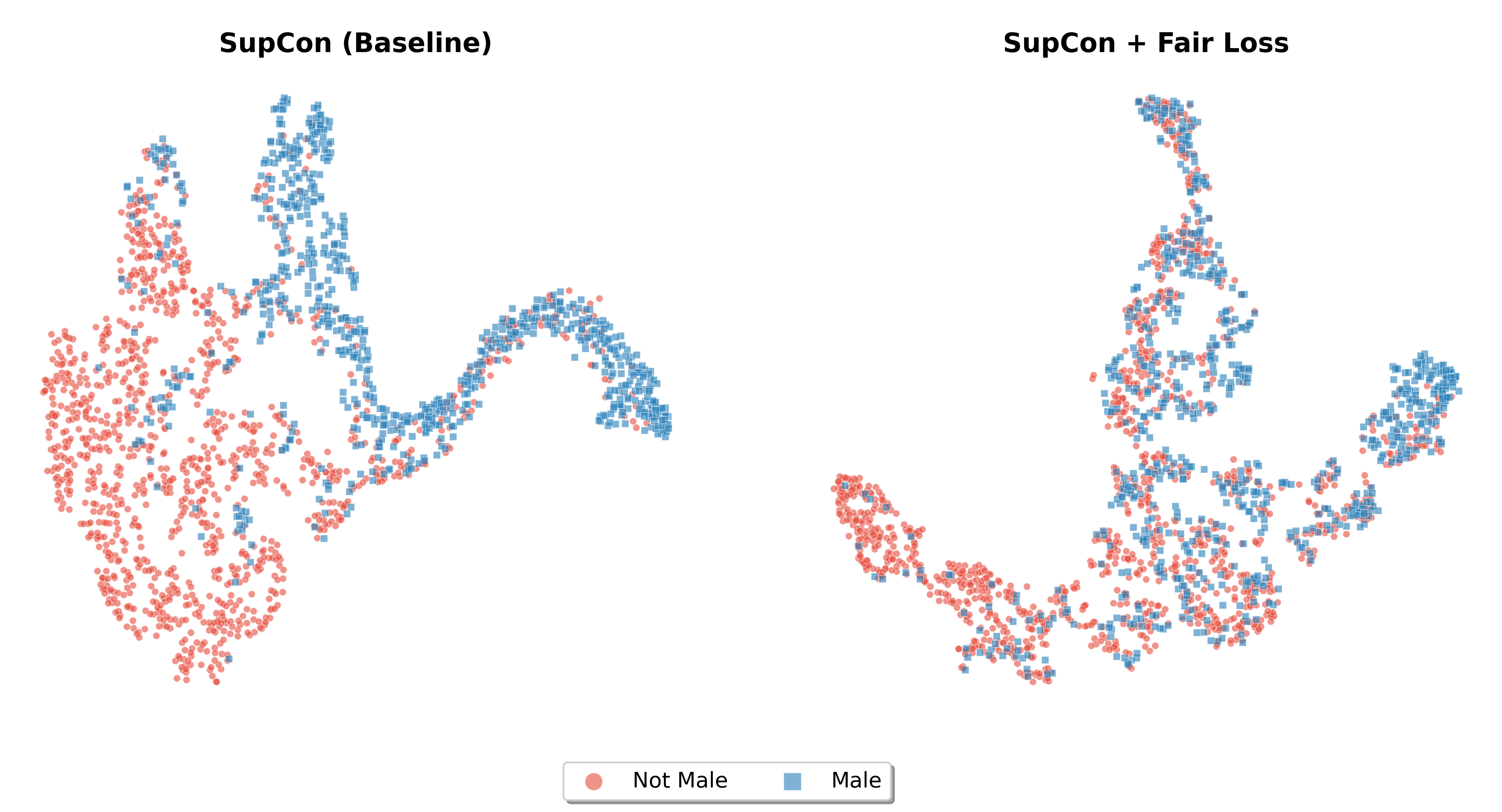}
      \caption{Target: \textit{Big Nose}}                                                                                     
      \label{fig:tsne-bignose}
    \end{subfigure}
    \hfill
    \begin{subfigure}{0.48\linewidth}
      \includegraphics[width=\linewidth]{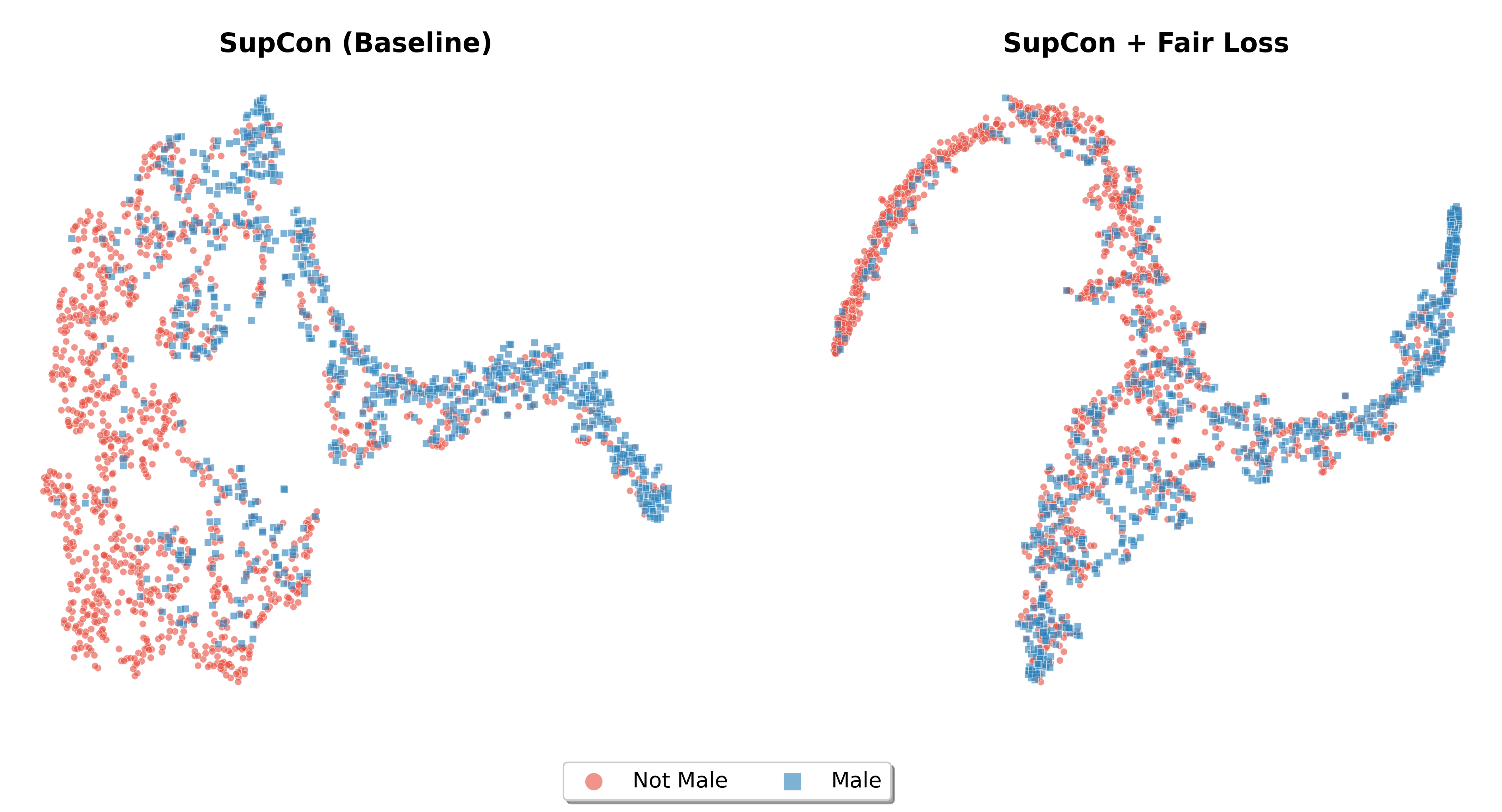}
      \caption{Target: \textit{Bags Under Eyes}}
      \label{fig:tsne-bags}
    \end{subfigure}
    \caption{t-SNE visualizations colored by the sensitive attribute (\textit{Male} vs.\ \textit{Not Male}). In each
  subfigure, baseline SupCon is shown on (\emph{left}) and SupCon + Fair Loss on (\emph{right}). 
  % The
  % fairness-regularized model produces representations with substantially greater overlap between the two sensitive groups
  % across both target tasks.
  }
    \label{fig:tsne-sensitive}
  \end{figure}

  \subsection{Qualitative Analysis of Representation Fairness}
To qualitatively assess whether ProtoFair mitigates the encoding of sensitive attribute information, we visualize the CelebA test-set embeddings using t-SNE~\cite{maaten2008visualizing}. We compare the baseline SupCon~\cite{khosla2020supervised} against our model (SupCon + ProtoFair) from Table~\ref{tab:celeba} on two target attributes, Big Nose and Bags Under Eyes, both with Male as the sensitive attribute, coloring points by the sensitive attribute (Male vs. Not Male). In the baseline embedding (Figure~\ref{fig:tsne-sensitive}(a), \emph{left}), Male and Not Male samples form largely disjoint clusters, indicating that the encoder has entangled gender with the target features. After applying ProtoFair for only 5 epochs (Figure~\ref{fig:tsne-sensitive}(a), \emph{right}), the two groups become substantially more intermixed, suggesting the regularizer reduces reliance on gender as a distinguishing feature. The same pattern holds for Bags Under Eyes (Figure~\ref{fig:tsne-sensitive}(b)). To corroborate these observations, we train a linear classifier on the frozen features to predict the sensitive attribute, where lower accuracy indicates less retained sensitive information. As shown in Table~\ref{tab:linprobe}, ProtoFair reduces sensitive-attribute predictability on both tasks, consistent with the t-SNE intermixing and the corresponding EO reductions.

\begin{table}[tb]
  \caption{Sensitive attribute predictability from frozen features on CelebA. Lower linear-probe accuracy indicates less sensitive information is retained. Sensitive attribute: Male.}
  \label{tab:linprobe}
  \centering
  % \scriptsize
  % \setlength{\tabcolsep}{5pt}
  \begin{tabular}{@{}lcccc@{}}
    \toprule
    & \multicolumn{2}{c}{Big Nose}
    & \multicolumn{2}{c}{Bags Under Eyes} \\
    \cmidrule(lr){2-3} \cmidrule(lr){4-5}
    Method & Probe-ACC $\downarrow$ & EO $\downarrow$& Probe-ACC $\downarrow$& EO $\downarrow$ \\
    \midrule
    SupCon ~\cite{khosla2020supervised}          & 87.76 & 20.7 & 82.41 & 20.8 \\
    SupCon + ProtoFair  & 81.02 & 15.8 & 76.07 & 10.1 \\
    \bottomrule
  \end{tabular}
\end{table}

\subsection{Ablation Study}
\begin{figure}[t]
  \centering
  % \fbox{\rule{0pt}{0.5in} \rule{0.9\linewidth}{0pt}}
\includegraphics[width=0.99\linewidth]{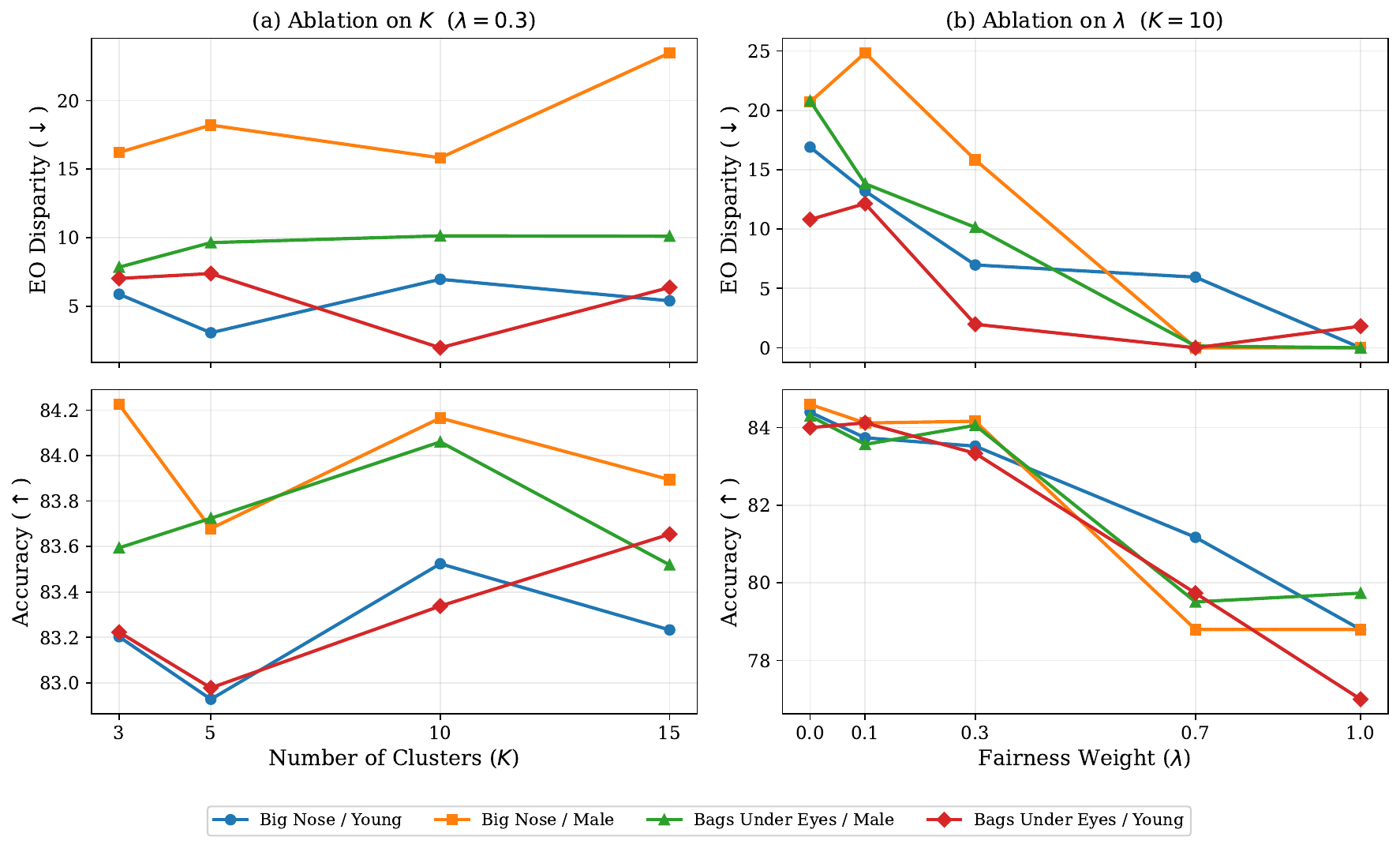}
\caption{Ablation on $K$ and $\lambda$. EO (top) and accuracy (bottom) on four CelebA (T, S) pairs, varying 
the number of clusters 
$K$ at $\lambda{=}0.3$ (left) and 
the fairness weight 
$\lambda$ at $K{=}10$ (right).}
   \label{fig:ablation}
\end{figure}

We study the sensitivity of ProtoFair to its two main hyperparameters: the number of prototypes $K$ and the fairness weight $\lambda$. Figure~\ref{fig:ablation} reports EO disparity and accuracy on four CelebA (T, S) pairs as we vary $K$ at $\lambda{=}0.3$ (left) and $\lambda$ at $K{=}10$ (right). Accuracy remains stable across $K$ (within 83.0--84.5), indicating that ProtoFair is robust to the cluster count. Varying $\lambda$ reveals a sweet spot in $[0.1, 0.3]$, where EO is reduced with little accuracy cost. Larger $\lambda$ values overconstrain the learned representation, leading to reduced accuracy when $\lambda \geq 0.7$ or higher. These trends support our default choice of $K{=}10$ and $\lambda \in [0.1, 0.3]$ used throughout our experiments.

\section{Discussion and Conclusions}
We introduced ProtoFair, a fairness-aware contrastive loss that complements existing self-supervised learning objectives without modifying them. The central idea is to leverage unsupervised clustering, which is typically used in self-supervised learning to enhance representation quality, in our fairness mechanism. Momentum-updated prototypes, initialized with K-Means and refined via exponential moving average, assign each sample to a cluster that serves as a proxy for its semantic content. Samples assigned to the same cluster but belonging to different sensitive groups form \emph{pseudo-counterfactual pairs}: they share content but differ in the sensitive attribute. Pulling these pairs together in the embedding space encourages the encoder to produce representations that are invariant to the sensitive attribute while preserving semantic content, without requiring target task labels. To the best of our knowledge, leveraging clustering based pseudo labels to construct fairness-aware contrastive    
  objectives, in particular, using them to identify cross-group counterfactual pairs has not been previously explored. 
% To the best of our knowledge, both the use of clustering-based pseudo-labels for fairness in contrastive learning and the proposed pair construction strategy are novel. 
A cross-batch queue further expands pair discovery beyond the current mini-batch, strengthening the fairness signal under group imbalance.

Experiments on CelebA, UTKFace, and NIH Chest X-rays show that ProtoFair consistently reduces equalized odds across supervised (SupCon) and self-supervised (SimCLR, BarlowTwins, BYOL) base methods, achieving fairness competitive with dedicated methods while maintaining higher accuracy. Results on UTKFace further confirms robustness across varying levels of data imbalance. While dedicated fairness architectures may reach stronger performance, ProtoFair offers a practical and modular alternative that meaningfully closes the fairness gap without requiring target label supervision.

% architectural changes or 

% A current limitation is that ProtoFair relies on the quality of cluster assignments to approximate content similarity. 

% ProtoFair is inherently dependent on the quality of cluster assignments, which serve as an approximation of content similarity. If clusters fail to capture meaningful semantic structure, the constructed pairs may not represent true counterfactuals. In practice, the warmup period and training schedule benefit from tuning per setting, since cluster quality depends on the maturity of the learned representations, which evolves at different rates depending on the base objective, dataset, and the degree of visual correlation between target and sensitive attributes. 
ProtoFair depends on the quality of cluster assignments, which approximate content similarity. If clusters fail to capture meaningful semantic structure, the resulting pairs may yield unreliable counterfactual approximations. We mitigate this by deriving clusters from \emph{learned} representations and scheduling the ProtoFair loss only after a warmup period, once the encoder has matured. 
% Since representation quality evolves at different rates across base objectives, datasets, and the degree of target/sensitive correlation, the warmup length still benefits from per-setting tuning.

Future work includes extending ProtoFair to safety-critical perception tasks such as autonomous driving, where pedestrian detection should not vary across demographic groups, and human-robot interaction, where social and assistive robots must perceive people equitably.
% Future work includes evaluating ProtoFair across a broader range of self-supervised paradigms beyond contrastive learning, such as Siamese methods (e.g., BYOL~\cite{BYOL}), generative prediction (e.g., GPT-style pretraining~\cite{Radford2018ImprovingLU}), and pretext task approaches (e.g., Jigsaw puzzles~\cite{DBLP}), to further validate its generality as a plug-in fairness loss. 

% Additionally, we plan to conduct systematic ablation studies over key hyperparameters, particularly the number of clusters $K$ and the fairness weight $\lambda$, to better understand their interaction with different sensitive attributes and dataset characteristics.

% \section*{Acknowledgements}
% Please insert your acknowledgments here.

\bibliographystyle{splncs04}
\bibliography{main}
\end{document}